\def\year{2021}\relax
\documentclass[letterpaper]{article} 
\usepackage{aaai21}  
\usepackage{times}  
\usepackage{helvet} 
\usepackage{courier}  
\usepackage[hyphens]{url}  
\usepackage{graphicx} 
\def\UrlFont{\rm}  
\usepackage{natbib}  
\usepackage{caption} 
\usepackage{subcaption}
\usepackage{multirow}
\usepackage{amsmath}
\usepackage[switch]{lineno}  %
\title{An Efficient 3D Convolutional Neural Network with Channel-wise, Spatial-grouped, and Temporal Convolutions}
\author{
    Zhe Wang, Xulei Yang\\
}
\affiliations{
    \{wang\_zhe, yang\_xulei\}@i2r.a-star.edu.sg, Institute for Infocomm Research, A*STAR, Singapore

}

\begin{document}
\maketitle

\begin{abstract}
There has been huge progress on video action recognition in recent years. However, many works focus on tweaking existing 2D backbones due to the reliance of ImageNet pretraining, which restrains the models from achieving higher efficiency for video recognition. In this work we introduce a simple and very efficient 3D convolutional neural network for video action recognition. The design of the building block consists of a channel-wise convolution, followed by a spatial group convolution, and finally a temporal convolution. We evaluate the performance and efficiency of our proposed network on several video action recognition datasets by directly training on the target dataset without relying on pertaining. On Something-Something-V1\&V2, Kinetics-400 and Multi-Moments in Time, our network can match or even surpass the performance of other models which are several times larger. On the fine-grained action recognition dataset FineGym, we beat the previous state-of-the-art accuracy achieved with 2-stream methods by more than 5\% using only RGB input.
\end{abstract}

\section{Introduction}
\begin{figure}[t]
    \centering\includegraphics[width=1\linewidth]{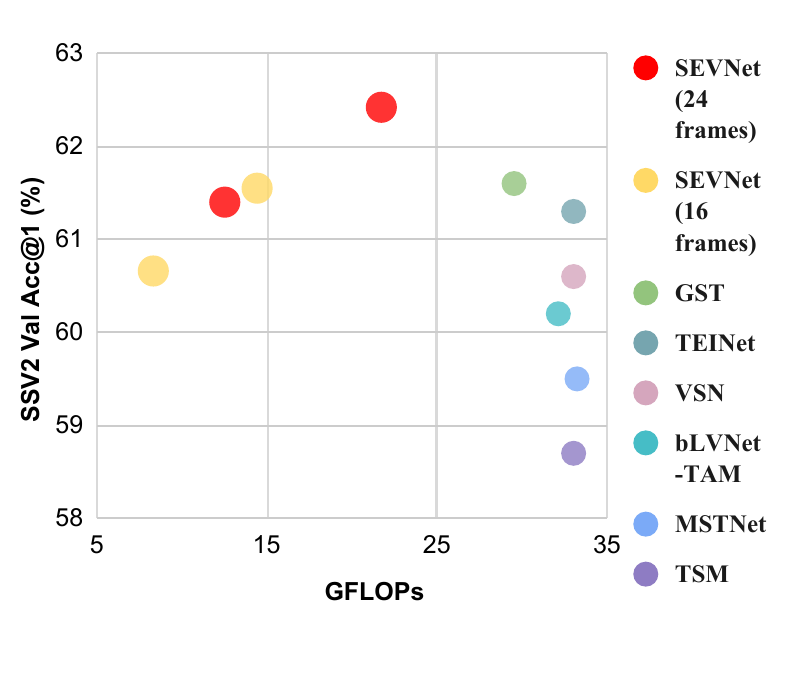}
    \caption{Accuracy against complexity (FLOPs) plot of state-of-the-arts models under 35 GFLOPs on Something-Something-V2.}
    \label{fig:SSV2_chart}
\end{figure}

Video action recognition is one of the fundamental research topics in computer vision. Since the introduction of ImageNet \cite{Deng2009CVPR} and AlexNet \cite{Krizhevsky2012NIPS}, deep convolutional neural network (CNN) has become the essential tool to solve many computer vision tasks. Over the last decade, the efficiency of image classification models has increased by around two order of magnitude through the development of works such as ResNet \cite{resnet}, MobileNet \cite{howard2017mobilenets} and EfficientNet \cite{tan2019efficientnet}. CNN has also been researched and utilized to solve the video action recognition task for several years. The early works \cite{Simonyan2014NIPS, Wang2016ECCV} started with using 2D networks to extract features on RGB frames and optical flow inputs. As videos contain an extra temporal dimension compared to images, research on using 3D CNN also emerged \cite{Tran2015ICCV,kinetics}. More recently, several works \cite{TIN,GST,TEINet,TSM,VSN} explored methods to modify existing 2D networks in order to improve their temporal modeling capability for action recognition.

Despite the great research progress, deep video models are still very heavy in terms of huge number of parameters and large computational cost. This makes it hard to deploy video recognition models for real world applications, especially those requiring real-time response under certain computational resource constraints. Therefore, improving the efficiency of video recognition models is an essential research problem to be studied. The inefficiency in existing methods is partly due to constraints of using 2D pretrained models. Recently, several works \cite{Feichtenhofer2019ICCV, Tran2019ICCV} have shown that with proper training strategy, models trained from scratch can achieve state-of-the-art accuracy on Kinetics-400 \cite{kinetics} dataset. However, a few relatively more recent works still employ pretraining when performing experiments on smaller datasets such as Something-Something V1 \cite{SSV1} \& V2 \cite{SSV2} and FineGym~\cite{finegym}. 

In this work, we show that it is actually possible to train 3D networks with simple structure from scratch to achieve state-of-the-art performance, with standard and consistent optimization parameters on different datasets. We introduce a simple residual block design by firstly performing channel-wise convolution, followed by a grouped spatial convolution, and finally a temporal convolution. By stacking these blocks repeatedly in the same fashion as ResNet-50~\cite{resnet}, we can build a series of very compact video recognition models named as \textbf{SEVNets}: \textbf{S}imple and \textbf{E}fficient \textbf{V}ideo \textbf{Net}works. 

We perform experiments on several popular video action classification datasets including Something-Something V1 \& V2, Multi-Moments in Time \cite{MMIT} and Kinetics-400 \cite{kinetics}. Although we directly train on the target datasets, our SEVNets can achieve similar or even better performance than related works which are several times larger. For example, as shown in figure \ref{fig:SSV2_chart}, our SEVNets are much more efficient with higher accuracy and lower computational complexity compared to other methods. Moreover, on the newly proposed fine-grained action classification dataset FineGym~\cite{finegym}, we beat the previous state-of-the-art accuracy achieved with 2-stream methods by more than 5\% using only RGB input. To the best of our knowledge, we are the first to achieve state-of-the-art performance on Something-Something V1 \& V2, Multi-Moments in Time and FineGym without any pretraining. 

Similar to one of the works that inspired us \cite{simple_pose} to construct a simple baseline for action recognition, we do not claim any superiority in terms of network design or training strategy over previous methods, despite having achieved similar or better performance with less computation. We show by experiments that  improvements to 2D/3D networks such as attention mechanism are orthogonal to our work, and therefore can be combined to achieve even better performance. The main contributions of this work are as follows:
\begin{itemize}
    \item A simple 3D convolutional network module is presented, and by stacking them we can build a series of efficient 3D networks named as SEVNets. The SEVNets produce solid baselines for future research in video action recognition.
    \item Comprehensive experiments on several popular video action recognition datasets show that SEVNets trained from scratch can match or even surpass the performance of other methods with pretraining on ImageNet and/or Kinetics. 
    \item The above results indicate that with a simple optimization strategy, pretraining is not necessary to achieve good performance for action classification, which stays true even on smaller datasets. We hope this can motivate further research in designing efficient 3D network structures for video action recognition, without being constrained to existing 2D structures.
\end{itemize}

\section{Related Works}
\textbf{2D CNNs in video action recognition}. Before introduction of 3D CNNs for video modeling, 2D CNNs were first used for action recognition by \cite{Simonyan2014NIPS, Wang2016ECCV}, where two separate networks were used for extracting features from RGB frames and motion information (optical flow) separately. Several other works \cite{feichtenhofer2016spatiotemporal,zhang2016real} followed similar approach with improvements in spatio-temporal modeling ability. However, it is usually expensive and troublesome to extract optical flow features, thus methods which require only single network emerged. Temporal Shift Module (TSM) \cite{TSM} shifts some of the channels along the temporal axis to achieve information mixing among neighboring frames, which replaces the separate network for motion feature extraction. A few other works \cite{TIN,GST,TEINet,VSN,Jiang2019ICCV} further explored different ways of incorporating and enhancing temporal modeling into 2D networks. While some of these methods have been successfully proven effective, most of them still rely on 2D networks (ResNet in most works) pretrained on ImageNet as a starting point, which limits their design space for greater efficiency.

\begin{figure}
  \begin{subfigure}[b] {.43\linewidth}
    \centering\includegraphics[width=.98\linewidth]{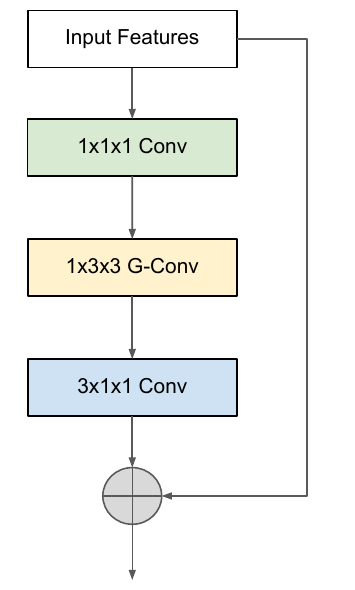}
    \caption{SEV Module}
    \label{fig:SEV_M}
  \end{subfigure}
  \hfill
  \begin{subfigure}[b] {.57\linewidth}
    \centering\includegraphics[width=.98\linewidth]{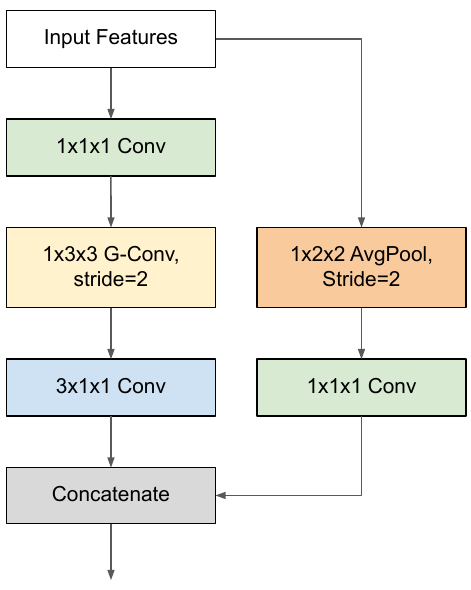}
    \caption{SEV Module (downsample)}
    \label{fig:SEV_D}
  \end{subfigure}
  \hfill
  \caption{Structure of SEV Module}
  \label{fig:SEV_B}
\end{figure}

\textbf{3D CNNs in video action recognition}. 
As there are 3 dimensions (2 spatial dimension and 1 temporal dimension) for video data, spatio-temporal 3D CNNs \cite{Tran2015ICCV} naturally become a suitable tool for video recognition. I3D \cite{kinetics} introduced an effective way to inflate 2D networks to 3D networks so that the parameters learned during pretraining on ImageNet can be leveraged for better generalization. In the past few years, several works \cite{qiu2017learning, r2p1d, Xie2018ECCV} tried to decouple one single 3D convolution into separate spatial convolution and temporal convolution, and explored different arrangements of combining them for better efficiency. Recently, depthwise spatio-temporal 3D convolutions have been utilized to increase the efficiency by reducing the computational complexity in CSN \cite{Tran2019ICCV}. SlowFast networks \cite{Feichtenhofer2019ICCV} proposed to employ two pathways for processing features at different frame rates, with lateral connections to fuse the features multiple times along the networks.

\textbf{On the usage of pretraining in related works}. Since the introduction of I3D \cite{kinetics}, most related works rely on ImageNet pretrained models when training on Kinetics-400, and ImageNet and/or Kinetics pretrained models when training on other action recognition datasets such as Something-Something V1 \cite{SSV1} \& V2 \cite{SSV2}. More recently, CSN \cite{Tran2019ICCV} and SlowFast \cite{Feichtenhofer2019ICCV} showed that with proper training strategies, it is not necessary to have ImageNet pretraining to achieve state-of-the-art results on Kinetics-400. However, whether it is possible to achieve competitive performance by training from scratch on other action recognition datasets (Something-Something V1 \& V2, Multi-Moments in Time, FineGym) remains unclear. 

\section{Simple and Efficient Video Networks}
In this section, we formally introduce the \textbf{S}imple and \textbf{E}fficient \textbf{V}ideo \textbf{Net}works (\textbf{SEVNets}). We first describe the motivation and structure of the building blocks of SEVNets, named as SEV Modules. Then we stack groups of SEV Modules in the same fashion as ResNet-50~\cite{resnet} to build SEVNets. Simplicity is the main principle of our design, therefore we do not include any complicated components such as multi-branching, channel shuffling or attention learning used by related works.

\subsection{SEV Module}
The design of the SEV Module is shown in figure \ref{fig:SEV_M}. The high level block design resembles the structure of a residual module~\cite{resnet}, where we add the features processed with convolutions with the input features for output. Inspired by R(2+1)D ~\cite{r2p1d}, we separate a single 3D spatial convolution into a spatial convolution followed by a temporal convolution. Furthermore, we adopt the more efficient group convolution for the spatial feature processing as it is relatively more costly due to having the largest kernel size within the block. We denote the group width as \textbf{G} which is a constant across different blocks throughout the whole network. Each convolution operation is followed by BatchNorm~\cite{Ioffe2015arXiv} and ReLU~\cite{Nair2010ICML} following common practice. For simplicity, the number of output channels for each convolution is the same as that for input. Though a bottleneck or inverted-bottleneck design with different number of channels in the middle might produce better results, it will require more parameter tuning which is not the focus of the paper. Moreover, as suggested by~\cite{regnet}, our block design with constant number of output channels is efficient especially on GPU.

Additionally, we shrink the spatial dimensions and increase the number of channels of the feature maps progressively in the network by utilizing SEV Module (downsample) as shown in figure \ref{fig:SEV_D}. We replace the identity path in the standard block by an average pooling layer followed by an $1\times1\times1$ convolution, and then concatenate with the main branch. This design pattern of the downsample block is inspired by ShuffleNet~\cite{shufflenet}. As both branches produce the same number of features as the input, the total number of channels for the output doubles as desired.

\subsection{SEVNets}

With the standard and downsample SEV modules, we can now construct SEVNets. The detailed structure of SEVNets is shown in table \ref{table:SEVNets}. The stem of the network consists of an $1\times7\times7$ spatial convolution with stride 2, followed by an $3\times1\times1$ temporal convolution. Then we apply 4 stages of SEV Modules one by one, the number of modules in each stage is (3,4,6,3) which follows the same configuration from ResNet-50. Though other configuration for number of modules in different stages might be superior, we just follow the standard ResNet-50 setting. This is because it would be too costly to search for the (near) optimal combination of parameters, which is not the focus of our work that targets at presenting a simple baseline. The first SEV Module in each stage is a downsample block, which is responsible for reducing the spatial dimension by half and doubling the number of channels. Similar to the width multiplier introduced in MobileNets~\cite{howard2017mobilenets}, \textbf{G} is the group width of all the group convolutions in the network, which is used to control the speed-accuracy trade-off. We denote the SEV Network with group width G=C by SEVNet-G\textbf{C}, i.e. SEVNet-G8 stands for the SEVNet with group width G=8.

    

\begin{table}[t]

    \centering
    \begin{tabular}{c|c|c}
    \hline
    \textbf{Stage}  & \textbf{Operation}  & \textbf{Output size}  \\
    \hline
    input & - & $3\times T\times H\times W$ \\
    \hline
    conv$_1$ & \begin{tabular}{@{}c@{}} $1\times7\times7$ (s=2), 4G  \\ $3\times1\times1$, 4G \end{tabular} & $4G\times T\times \frac{H}{2}\times \frac{W}{2}$  \\
    \hline
    $\text{stage}_2$ & \(\begin{bmatrix}  1\times1\times1, 8G \\ 1\times3\times3, 8G \\ 3\times1\times1, 8G \end{bmatrix} \times 3\)  & $8G\times T\times \frac{H}{4}\times \frac{W}{4}$  \\
    \hline
    $\text{stage}_3$ & \(\begin{bmatrix}  1\times1\times1, 16G \\ 1\times3\times3, 16G \\ 3\times1\times1, 16G \end{bmatrix} \times 4\)  & $16G\times T\times \frac{H}{8}\times \frac{W}{8}$  \\
    \hline
    $\text{stage}_4$ & \(\begin{bmatrix}  1\times1\times1, 32G \\ 1\times3\times3, 32G \\ 3\times1\times1, 32G \end{bmatrix} \times 6\)  & $32G\times T\times \frac{H}{16}\times \frac{W}{16}$  \\
    \hline
    $\text{stage}_5$ & \(\begin{bmatrix}  1\times1\times1, 64G \\ 1\times3\times3, 64G \\ 3\times1\times1, 64G \end{bmatrix} \times 3\)  & $64G\times T\times \frac{H}{32}\times \frac{W}{32}$  \\
    \hline
    pool & Global Avg Pool & $64G\times 1\times1\times1$ \\
    \hline
    fc & FC &  \# classes \\
    \hline
    \end{tabular}
    \caption{
        \textbf{\textbf{S}imple and \textbf{E}fficient \textbf{V}ideo \textbf{Net}works (\textbf{SEVNets})}. G stands for the group width of all group convolutions.
    }
    \label{table:SEVNets}
\end{table}

\section{Experiments}

\begin{table*}[t]
\centering
    \begin{center}
        
        \begin{tabular}{l|l|l|l|c|c}
        \hline
        Model & Pretrain  & M.Params & GFLOPs & SSV1 Acc@1(\%) & SSV2 Acc@1(\%) \\
        \hline\hline
        TIN~\cite{TIN}                 & Kinetics & 24.3 & 34 & 45.8 & - \\
        TSM~\cite{TSM}         & ImageNet & 24.3 & 33 & 45.6 & 58.7 \\
        STM~\cite{Jiang2019ICCV}       & ImageNet & 24.0 & 33.3 & 47.5 & - \\
        MSTNet~\cite{MixTConv}         & ImageNet & 24.3 & 33.2 & 46.7 & 59.5 \\
        bLVNet-TAM~\cite{fan2019more}  & ImageNet & 40.2 & 32.1 & 47.8 & 60.2 \\
        VSN~\cite{VSN} & ImageNet & 24.3 & 33 & 46.6 & 60.6 \\
        TEINet~\cite{TEINet}     & ImageNet & 30.4 & 33 & 47.4 & 61.3 \\
        GST~\cite{GST}         & ImageNet & 21.0 & 29.5 & 47.0 & 61.6 \\

        \hline
        \textbf{SEVNet-G6 (16 frames)} & - & \textbf{2.5} & \textbf{8.3} & 47.6 & 60.7 \\
        \textbf{SEVNet-G8 (16 frames)} & - & \textbf{4.4} & \textbf{14.4 }& \textbf{48.5} & 61.6 \\
        \hline
        \textbf{SEVNet-G6 (24 frames)} & - & \textbf{2.5} & \textbf{12.5} & \textbf{48.4} & 61.4 \\
        \textbf{SEVNet-G8 (24 frames)} & - & \textbf{4.4} & \textbf{21.7} & \textbf{49.1} & \textbf{62.4} \\
        \hline

        \end{tabular}
        \caption{\textbf{Comparison of SEVNets with models under 35 GFLOPs on Something-Something-V1\&V2.} 
        }
        \label{table:ss}
    \end{center}
\end{table*}

In this section we first describe the details of training and testing procedures of the experiments. Then we evaluate the effectiveness of our SEVNets against current state-of-the-art methods on several video action recognition datasets including Something-Something V1 \cite{SSV1} \& V2 \cite{SSV2}, Multi-Moments in Time \cite{MMIT} and Kinetics-400 \cite{kinetics}. 
Additionally, we perform experiments on the recently proposed fine-grained action classification dataset FineGym~\cite{finegym} and report new state-of-the-art performance with more than 5\% improvements despite using only RGB input. We use PyTorch \cite{paszke2019pytorch} for our experiments.

\subsection{Training and Testing}\label{train_test}

The following is the common training and testing setup for most of the experiments that will be presented. Additional specific dataset related settings (if any) are described in the corresponding subsections. For all the experiments, we only take RGB channels of the video frames as input. No optical flow information is used as it is costly to obtain and will double the complexity of the whole model.

\textbf{Training}. Each model is trained for a total of 64 epochs on 8 GPUs with 8 samples on each GPU, making the total batch-size 64. We set the base learning rate as 0.01 per GPU following \cite{Tran2019ICCV}. A cosine learning rate schedule \cite{Loshchilov2016arXiv} is used with the first 4 epochs as the warm-up period. All models are \textbf{trained from scratch} and are optimized using Stochastic Gradient Descent with momentum at 0.9 and weight decay of 0.0001. To avoid overfitting, we make use of a Dropout \cite{dropout} layer before the classification layer. We set 0.5 as the dropout rate for the relatively small and medium datasets (FineGym and Something-Something V1 \& V2), and 0.2 for the larger datasets (Kinetics-400 and Multi-Moments in Time). Following the sparse segment-based sampling method proposed by \cite{Wang2016ECCV}, we uniformly divide the video into 16 segments, and randomly sample one frame from each segment to form a video clip of 16 frames. For spatial sampling, we first resize the length of the shorter side to a number uniformly sampled from the range of 224 to 256 while keeping the original aspect ratio. We then randomly crop a squared region with size 224 from the resized frames as the input to the network during training. The input size to the models here in terms of $T\times H\times W$ is thus $16\times 224 \times 224$.

\textbf{Testing}. For inference, we select the middle frame from each of the 16 uniformly divided segments. All frames are re-scaled such that the length of the shorter side is equal to 224. We then perform one single center crop to obtain a squared region of $224\times 224$ from the resized frames. Finally, we evaluate the single-crop performance against other related works on the validation set of respective datasets.

\begin{figure}
    \centering\includegraphics[width=1\linewidth]{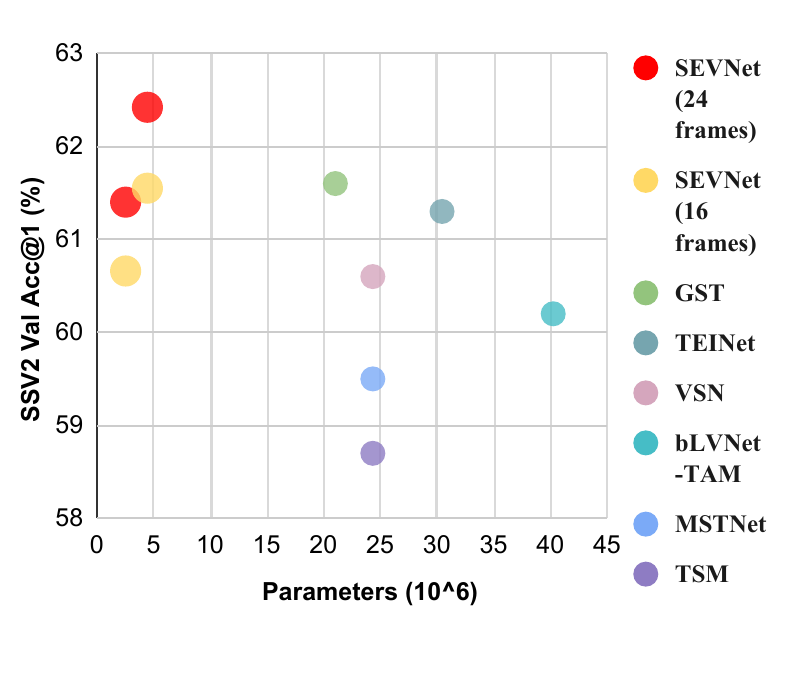}
    \caption{Accuracy against number of parameters plot of state-of-the-arts models under 35 GFLOPs on Something-Something-V2.}
    \label{fig:SSV2_param}
\end{figure}

\subsection{Experiments: Something-Something V1 \& V2}

\textbf{Something-Something-V1 (SSV1)} \cite{SSV1} is an action recognition datasets that shows humans performing certain actions with some objects. The total number of action categories is 174. There are around 86k and 11.5k video clips in the training and validation set respectively. \textbf{Something-Something-V2 (SSV2)} \cite{SSV2} doubled the number of videos, improved the annotation quality and increased the resolution of videos compare to V1. These two versions of the dataset are known to be temporal-sensitive in the way that they require understanding the action more than the appearance of background or objects. Hence they are more suitable for assessing the spatio-temporal modeling capabilities.

Table \ref{table:ss} shows the performance comparison of our SEVNets with other recently proposed state-of-the-art methods on the validation set of SSV1 and SSV2 with single-crop testing. Note that all other models in the table are pretrained on either ImageNet or Kinetics-400, while SEVNets are \textbf{trained from scratch} directly on SSV1 and SSV2. Among all the "small" version (under 35 GFLOPs) of other methods, the previous state-of-the-art accuracy for SSV1 and SSV2 were 47.8\% and 61.6\% respectively. Our SEVNet-G8 (16 frames) achieved comparable and higher accuracy of 61.6\% for SSV2 and 48.5\% for SSV1, while using only less than half of computation cost and less than $1/5$ of parameters. Moreover, as we know that the SSV1 and SSV2 datasets require more temporal modeling ability, we perform additional experiments using 24 frames as input instead of 16. While still satisfying the constraints of 35 GFLOPs of computation, SEVNet-G8 (24 frames) achieved even higher accuracy of 49.1\% and 62.4\% for SSV1 and SSV2 respectively.

Figure \ref{fig:SSV2_chart} and figure \ref{fig:SSV2_param} are visual illustrations of accuracy against computational complexity and number of parameters respectively, which includes several state-of-the-art methods on the validation set of SSV2. We can observe that our SEVNets can match and even outperform other methods while using only a fraction of the computational cost as well as parameters.

\subsection{Experiments: Multi-Moments in Time}

\begin{table}
    \begin{center}

        \begin{tabular}{l|l|c|c}
        \hline
        Model & Pretrain  & FLOPs & mAP\% \\
        \hline\hline
        
        \begin{tabular}{@{}l@{}}TSN\\ \cite{Wang2016ECCV}\end{tabular}   & K400 &  \textasciitilde 40G & 58.92 \\
        \hline
        \begin{tabular}{@{}l@{}}TSM\\ \cite{TSM}\end{tabular}   & K400 &  \textasciitilde 63G & 61.06 \\
        \hline
        \begin{tabular}{@{}l@{}}TIN\\ \cite{TIN}\end{tabular}    & K400 &  \textasciitilde 64G & 62.22 \\ 
        
        \hline\hline
        \textbf{SEVNet-G8 (ours)} & - & \textbf{14.4G}  & 59.07 \\
        \textbf{SEVNet-G12 (ours)} & - & \textbf{31.8G}  & \textbf{62.33} \\
        
        \hline
        \end{tabular}
        \caption{\textbf{Comparison of SEVNets with state-of-the-arts on MMIT.} The mAP numbers for other methods are referenced from the Temporal Interlacing Network \cite{TIN}, while the FLOPs are estimated based on the information about the ResNet-101 backbone.}
        \label{table:mmit}        
        
    \end{center}
\end{table}

\textbf{Multi-Moments in Time (MMIT)} \cite{MMIT} is a recently published large-scale and multi-label video recognition dataset. It contains around 1.02 million videos with 2.01 million labels from 313 action categories. All the videos are trimmed to 3 seconds long. Similar to the Something-Something datasets, MMIT also put more emphasis on temporal-relationships for determining the action than the visual appearance compared to Kinetics-400. The main evaluation metric for MMIT is mean Average Precision (mAP).

Table \ref{table:mmit} shows the results of our SEVNets compared to other related works. The SEVNet-G12 obtained higher mAP of 62.33\%, but only cost less than half of the computational complexity of TIN \cite{TIN}. To the best of our knowledge, this is the first time on MMIT that a model can reach such high performance by directly training on the target dataset, without relying on pretraining with Kinetics-400 which was adopted by previous works.

\subsection{Experiments: Kinetics-400}

\begin{table*}[t]
    \begin{center}
        \begin{tabular}{l|l|c|c|c|c|c}
        \hline
        Model & Pretrain  & M.Params & \#views  & GFLOPs$^*$ & Acc@1(\%) & Acc@5(\%) \\
        \hline\hline
        ip-CSN~\cite{Tran2019ICCV}     &  -       & 24.5 & 10x1 & 15.9 & 71.8 & - \\
        CorrNet~\cite{corrnet}     &  -       & - & 10x1 & 37.4 & 73.4 & - \\
        bLVNet-TAM~\cite{fan2019more}  & ImageNet & 25   & 3x3  & 93.4 & 73.5 & 91.2 \\
        STM~\cite{Jiang2019ICCV}       & ImageNet & 24.0 & 10x3 & 66.5 & 73.7 & 91.6 \\
        STH~\cite{sth}       & ImageNet & 23.2 & 10x3 & 61 & 74.1 & 91.6 \\
        S3D-G~\cite{Xie2018ECCV}         & ImageNet & 11.6 & 10x3 & 71.4 & 74.7 & \textbf{93.4} \\
        TSM~\cite{TSM}     & ImageNet & 24.3 & 10x1$^+$ & 65   & 74.7 & - \\
        TEINet~\cite{TEINet}     & ImageNet & 31   & 10x3 & 33   & 74.9 & 91.8 \\
        TEA~\cite{tea}     & ImageNet & -   & 10x3 & 35   & 75.0 & 91.8 \\
        SlowFast~\cite{Feichtenhofer2019ICCV}  & - & 34.4   & 10x3 & 36.1 & \textbf{75.6} & 92.1 \\
        
        \hline
        SEVNet-G8 (ours) & - & \textbf{4.5} & 10x3 & \textbf{14.4} & 72.4 & 90.9 \\
        SEVNet-G12 (ours) & - & \textbf{10.0} & 10x3 & 31.8 & 75.0 & 91.9 \\

        \hline
        \multicolumn{7}{l}{$^*$ Complexity for single-crop inference. $^+$ Fully convolutional testing without cropping.}
        
        \end{tabular}
        \caption{\textbf{Comparison of SEVNets with other methods on Kinetics-400.}}
        \label{table:kinetics}        
        
    \end{center}
\end{table*}

\textbf{Kinetics-400} \cite{kinetics} is a large-scale video action recognition dataset, which contains 400 human action categories, with each having at least 400 video clips. All the clips are from YouTube videos and are trimmed to 10 seconds in duration. For Kinetics-400, the appearance of the background contributes greatly to the process of identifying the action category in the video. Therefore pretraining on ImageNet \cite{Deng2009CVPR} is generally considered beneficial especially for the Kinetics-400 dataset as compared to other datasets. However, recent developments of SlowFast Networks \cite{Feichtenhofer2019ICCV} and CSN \cite{Tran2019ICCV} have shown that with proper training strategy, it is possible to achieve similar or even better results without pretraining. We train our SEVNets by training from scratch for 100 epochs on this dataset, and our results again confirms that pretraining on ImageNet is not necessary. 

When experimenting with the Kinetics-400 dataset, a different temporal sampling method is used as the raw videos are longer (around 10 seconds). Following the frame sampling strategy in \cite{TEINet}, we uniformly sample 16 frames from a trimmed clip formed by 64 consecutive raw frames from the original video. We randomly sample one trimmed clip from a video during training, and uniformly sample 10 such clips for testing following common practice. For evaluation, we apply the standard 3 spatial crops for each trimmed clip, then average the predictions from the 10x3 views. We measure the top-1 and top-5 accuracy on the validation set for comparison with other methods.

Table \ref{table:kinetics} shows the comparison of our models to other related works. We listed versions of models that use only RGB frames as input and are not too much larger than SEVNets for fair comparison. Despite not having pretraining on ImageNet, our SEVNets still achieve comparable results as other models while requiring fewer computations and parameters.

\subsection{Experiments: Fine-grained Action Recognition on FineGym}
\begin{table*}[t]
    \begin{center}
        \begin{tabular}{l|l|c|c|cc|cc}
        \hline
        \multirow{2}{*}{ver.$^*$} & \multirow{2}{*}{Model}  & \multirow{2}{*}{Modality} &  \multirow{2}{*}{GFLOPs} & \multicolumn{2}{c}{Gym288} & \multicolumn{2}{c}{Gym99} \\
        ~&~&~&~& Mean & Top-1 & Mean & Top-1 \\
        \hline\hline
        \multirow{10}{*}{v1.0} & \multirow{2}{*}{TSN \cite{Wang2016ECCV}} & RGB 
        & -  & 26.5 & 68.3 & 61.4 & 74.8 \\
        ~&~& RGB+Flow & - & 37.6 & 79.9 & 76.4 & 86.0 \\
        \cline{2-8}
        
        ~ & I3D \cite{kinetics} & RGB & - & 28.2 & 66.1 & 64.4 & 75.6 \\
        \cline{2-8}
        ~ & NL I3D \cite{Wang2018CVPR} & RGB & - & 28.0 & 67.0 & 64.3 & 75.3 \\
        \cline{2-8}        
        
        ~ & \multirow{2}{*}{TRN \cite{trn}} & RGB 
        & - & 33.1 & 73.7 & 68.7 & 79.9 \\        
        ~ & ~  & RGB+Flow & - & 42.9 & 81.6 & 79.8 & 87.4 \\
        \cline{2-8}
        
        ~ & \multirow{2}{*}{TRNms \cite{trn}} & RGB 
        & - & 32.0 & 73.1 & 68.8 & 79.5 \\        
        ~ & ~  & RGB+Flow &  - & 43.3 & 82.0 & 80.2 & 87.8 \\
        \cline{2-8}

        ~ & \multirow{2}{*}{TSM \cite{TSM}} & RGB 
        & - & 34.8 & 73.5 & 70.6 & 80.4 \\        
        ~ & ~  & RGB+Flow & - & 46.5 & 83.1 & 81.2 & 88.4 \\
        \cline{2-8}

        \hline
        \hline
        v1.0 & SEVNet-G8 (ours) & RGB & \textbf{14.4} & \textbf{51.8} & \textbf{85.8} & \textbf{87.1} & \textbf{91.2} \\
                  
        \hline
        v1.1 & SEVNet-G8 (ours) & RGB & \textbf{14.4} & \textbf{58.5} & \textbf{87.9} & \textbf{88.4} & \textbf{91.9} \\       
 
        \hline
        \multicolumn{8}{l}{$^*$ Version of the training set used.}
        
        \end{tabular}
        \caption{\textbf{Comparison of SEVNets with state-of-the-arts on FineGym.}
        The accuracy numbers for all the other methods are directly referenced from \cite{finegym}.}. 
        \label{table:finegym}        
        
    \end{center}
\end{table*}

FineGym~\cite{finegym} is a recently proposed dataset for fine-grained action recognition, which is built with gymnastic videos. Comparing with other datasets, it contains finely defined labels for sub-actions of a set of gymnastic events. The actions that belong to the same gymnastic event often only have very subtle differences, and they usually have very similar or even the same background and camera angle. These difficulties pose great challenges to existing action recognition models. There were two fine-grained sub-tasks introduced in FineGym, \textit{Gym288} and \textit{Gym99} which contain 288 and 99 fine-grained action categories respectively. \textit{Gym288} contains 22671/9646 samples under the train/val set respectively, which means it follows a long-tailed distribution with less than 100 samples per class on average. By removing the action classes with too few samples from \textit{Gym288}, \textit{Gym99} is a more balanced subset containing 20484/8521 samples under the train/val set respectively. There are two evaluation metrics used for the fine-grained action classification tasks. The first is the standard \textit{Top-1 accuracy}, which is the fraction of correctly predicted samples over the total number of samples in the validation set. However due to the unbalanced distribution, it might overfit to the more frequent action classes. Therefore, the \textit{Mean accuracy} is proposed for a less biased view of the performance by calculating the average of per-class accuracy.

Same as other datasets, we train our SEVNets on the training set directly from scratch with the same protocol introduced in \ref{train_test}, and without using any special techniques for fine-grained recognition. Table \ref{table:finegym} shows the results of our SEVNets compared to other existing methods on \textit{Gym288} and \textit{Gym99}. Note that models for other methods were pretrained on ImageNet and/or Kinetics-400, though it might not be helpful for the action recognition task on FineGym. The specific computational complexities for the other methods were not explicitly mentioned in \cite{finegym}, but we estimate that they are at the level of at least around 50 GFLOPs for single modality (assuming ResNet-50 as backbone, and an input size of $12\times224\times224$). Our SEVNet-G8 only costs 14.4 GFLOPs of computation which is less than 40\% of the complexity of others. On \textit{Gym288}, we improve the state-of-the-art mean accuracy from 46.5\% obtained by the two-stream TSM \cite{TSM} to 51.8\% with only RGB input. Similarly, we improve the state-of-the-art mean accuracy on \textit{Gym99} from 81.2\% to 87.1\%. The authors of FineGym suggested that optical flow information and temporal dynamics are critical in achieving good results on these fine-grained action recognition tasks. Despite this, SEVNets can still achieve \textbf{more than 5\% absolute increase} in mean accuracy compared to two-stream models (and \textbf{more than 16\%} compared to models using only RGB input), which shows that our SEVNets possess great spatial-temporal modelling ability.

Furthermore, the authors of FineGym have introduced an updated version 1.1 for the dataset with more samples in the training set. Therefore we also perform experiments by training on the updated v1.1 train set, and test on the same validation set. The results are listed in the last row of table \ref{table:finegym} as a reference for future works. As expected, the mean accuracy for \textit{Gym288} and \textit{Gym99} are further improved to 58.5\% and 88.4\% respectively .

\subsection{Experiments: Ablation study}

\begin{figure}[t]
  \begin{subfigure}[b] {.48\linewidth}
    \centering\includegraphics[width=.9\linewidth]{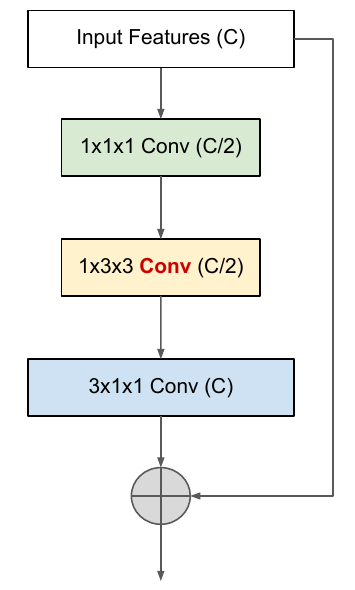}
    \caption{"R(2+1)D"}
    \label{fig:SEV_R2P1D}
  \end{subfigure}
  \begin{subfigure}[b] {.48\linewidth}
    \centering\includegraphics[width=.9\linewidth]{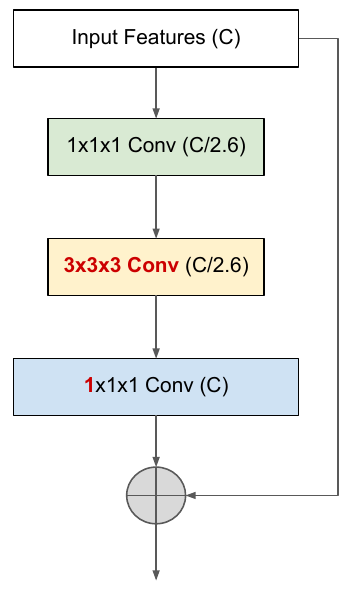}
    \caption{"R3D"}
    \label{fig:SEV_R}
  \end{subfigure}

  \caption{"R(2+1)D" and "R3D" like structures for ablation. (Numbers in brackets indicate number of output channels)}
  \label{fig:SEV_AB}
\end{figure}

In this subsection, we conduct experiments to verify the effectiveness of our design. The main novelty of the SEV module is the usage of grouped spatial convolution, which is not present in other related works in video recognition. By replacing the grouped spatial convolution to a normal convolution, the module is reduced to something similar to the structure of R(2+1)D \cite{r2p1d}, as shown in figure \ref{fig:SEV_R2P1D}. If we further combine the spatial and temporal convolutions to one single 3D spatio-temporal convolution, then the structure shown in \ref{fig:SEV_R} is similar to R3D \cite{r2p1d}. Note that, as we are performing ablation study with these two structures, we shrink the number of channels in the middle by 2x and 2.6x of the input respectively, such that the number of parameters and computational complexity are at comparable levels to the SEV module proposed by us. We build two series of networks by replacing the SEV Module (figure \ref{fig:SEV_M}) with "R(2+1)D" and "R3D" modules respectively for ablation study, with other parameters unchanged.

As shown in table \ref{table:ablate}, our proposed design with spatial group convolution consistently outperforms the version with standard spatial convolution ("R(2+1)D") by at least 1\% for models of similar sizes. When comparing to the "R3D" structure as the building block for the network, the improvement is even higher at more than 2\%.

\begin{table}
    \begin{center}
        \begin{tabular}{l|c|c|c|c}
        \hline
        \begin{tabular}{@{}l@{}}Block \\ structure \end{tabular} & G & \begin{tabular}{@{}c@{}}Million \\Params \end{tabular} & GFLOPs & \begin{tabular}{@{}c@{}}SSV2 \\Acc@1(\%)\end{tabular}   \\
        \hline\hline
        "R3D" & 6 & 2.8 & 8.4  & 57.57 \\
        "R3D" & 8 & 4.9 & 14.7 & 59.38 \\
        \hline
        "R(2+1)D" & 6 & 2.5 & 7.7  & 59.32 \\
        "R(2+1)D" & 8 & 4.4 & 13.4 & 60.51 \\
        \hline
        SEV (ours) & 6 & 2.5 & 8.3 & \textbf{60.66} \\
        SEV (ours) & 8 & 4.4 & 14.4 & \textbf{61.55} \\
        \hline
        \end{tabular}
        
        \caption{\textbf{Ablation experiments for SEV Module}}
        \label{table:ablate}
    \end{center}
\end{table}

\subsection{Experiments: Compatibility}
As we are proposing a simple baseline for action recognition, our model is compatible with other orthogonal methods that improve the performance of 2D/3D networks, such as attention mechanism. The Squeeze-and-Excitation (SE) \cite{senet} is one of the most popular methods for boosting performance of an existing backbone. Therefore we would like to verify the compatibility of SEVNets by running a simple set of experiments with it. We extend our SEVNets with SE units (with reduction ratio at 4) by applying them to the residual before the element wise addition in every SEV module. As table \ref{table:se} shows, the inclusion of SE units consistently improves the accuracy with slightly more parameters and negligible increase in computation. 

\subsection{Discussions: Complexity}
Deep Neural Networks typically have high computational complexity \cite{cai2019maxpoolnms, wang2017truly, yuan2017end}.
To reduce the time costs and power consumptions, some compression algorithms can be applied, including quantization \cite{young2021transform, zhe2019optimizing, zhe2021rate, wang2022rdo, chen2021efficient}, pruning \cite{xu2024lpvit, xu2023efficient}, feature coding \cite{duan2015weighted, wang2015hamming, wang2016affinity, duan2018minimizing, wang2016project, wang2014component, chen2012multi, wang2015optimizing, wu2016compact, lin2015selective}, etc.
We have tried different compression approaches to reduce the size and complexity of the model. Based on our results, typically the weights and activations of the model can be quantized to 8 bits without hurting the accuracy, which means about 4x compression compared with the original model.

\section{Conclusion}
In this work, we present a simple and efficient baseline for video action recognition. Our SEVNets, while being much more compact in terms of computation complexity and number of parameters, can achieve similar or even better performance than existing arts on a few popular action recognition datasets. On the fine-grained action recognition dataset FineGym, SEVNets with RGB-only input improves the state-of-the-art accuracy by more than 5\% compared to heavy two-stream models. Through our comprehensive experiments on various datasets, we show that it is possible to achieve comparable performance by directly training on the target dataset without any pretraining, not only on Kinetics, but also on other smaller datasets such as Something-Something-V1 \& V2. We hope this work can motivate further exploration of designing efficient 3D structures for video action recognition, instead of constraining research in tweaking existing 2D networks.

\begin{table}[t]
    \begin{center}
        \begin{tabular}{l|c|c|c|c}
        \hline
        Structure & G & \begin{tabular}{@{}c@{}}Million \\Params \end{tabular} & GFLOPs & \begin{tabular}{@{}c@{}}SSV2 \\Acc@1(\%)\end{tabular}   \\
        \hline\hline
        SEVNet & 6 & 2.5 & 7.7 & 60.66 \\
        SEVNet & 8 & 4.4 & 13.4 & 61.55 \\
        \hline
        SEVNet+SE & 6 & 2.8 & 7.7 & 61.65 \\
        SEVNet+SE & 8 & 4.9 & 13.4 & 62.47 \\
        \hline
        \end{tabular}
        
        \caption{\textbf{Experiments with SE unit}}
        \label{table:se}
    \end{center}
\end{table}

\bibliography{references}
\end{document}